\documentclass[12pt]{article}
\usepackage{amsmath}
\usepackage{graphicx,psfrag,epsf}
\usepackage{enumerate}
\usepackage{natbib}
\usepackage{float}
\usepackage{tablefootnote}
\usepackage{url} 
\usepackage{algorithm2e}
\usepackage{color}
\usepackage{multirow}
\usepackage{xr}
\makeatletter
\newcommand*{\addFileDependency}[1]{
  \typeout{(#1)}
  \@addtofilelist{#1}
  \IfFileExists{#1}{}{\typeout{No file #1.}}
}
\makeatother

\newcommand*{\myexternaldocument}[1]{%
    \externaldocument{#1}%
    \addFileDependency{#1.tex}%
    \addFileDependency{#1.aux}%
}

\myexternaldocument{SupplementaryDocument}

\newcommand{\blind}{0}

\newcommand{\matern}{Mat\'{e}rn}
\newcommand{\ba}{\ensuremath{\mathbf{a}}}

\newcommand{\bdelta}{\ensuremath{\boldsymbol{\delta}}}

\newcommand{\bepsilon}{\ensuremath{\boldsymbol{\epsilon}}}

\newcommand{\blambda}{\boldsymbol{\lambda}}

\newcommand{\boldeta}{\ensuremath{\boldsymbol{\eta}}}

\newcommand{\btheta}{\ensuremath{\boldsymbol{\theta}}}
\newcommand{\bu}{\ensuremath{\mathbf{u}}}
\newcommand{\bh}{\ensuremath{\mathbf{h}}}
\newcommand{\bc}{\ensuremath{\mathbf{c}}}
\newcommand{\bx}{\ensuremath{\mathbf{x}}}

\newcommand{\br}{\ensuremath{\mathbf{r}}}
\newcommand{\bw}{\ensuremath{\mathbf{w}}}
\newcommand{\bW}{\ensuremath{\mathbf{W}}}

\newcommand{\bY}{\ensuremath{\mathbf{Y}}}

\newcommand{\bZ}{\ensuremath{\mathbf{Z}}}

\begin{document}

\def\spacingset#1{\renewcommand{\baselinestretch}%
{#1}\small\normalsize} \spacingset{1}


\if0\blind
{
  \title{\bf \large Computer Model Calibration with Time Series Data \\ using Deep Learning and Quantile Regression}
  \author{Saumya Bhatnagar\hspace{.2cm}\\
    Division of Statistics and Data Science,\\ University of Cincinnati, Cincinnati, OH 45221-0025\\
    ~ \\
    Won Chang \thanks{
    Saumya Bhatnagar is Doctoral Candidate, Division of Statistics and Data Science, University of Cincinnati, OH 45221 (E-mail: bhatnasa@mail.uc.edu). Won Chang is Assistant Professor, Division of Statistics and Data Science, University of Cincinnati, OH 45236 (E-mail: changwn@ucmail.uc.edu). Seonjin Kim is Associate Professor, Department of Statistics, Miami University, OH 45056 (E-mail: kims20@miamioh.edu). Jiali Wang is Assistant Atmospheric Scientist, Argonne National Laboratory, IL 60439.}\\
    Division of Statistics and Data Science,\\ University of Cincinnati, Cincinnati, OH 45221-0025\\
    ~ \\
    Seonjim Kim\hspace{.2cm}\\
    Department of Statistics,\\ Miami University, Oxford, OH 45056\\
    ~ \\
    Jiali Wang\\
    Division of Environmental Science\\
    Argonne National Laboratory, Lemont, IL 60439}
  \maketitle
} \fi

\if1\blind
{
  ~\bigskip
  ~\bigskip
  \begin{center}
    {\large \bf Computer Model Calibration with Time Series Data \\ using Deep Learning and Quantile Regression}
\end{center}
  \medskip
} \fi

\newpage

\begin{abstract}
Computer models play a key role in many scientific and engineering problems. One major source of uncertainty in computer model experiment is input parameter uncertainty. Computer model calibration is a formal statistical procedure to infer input parameters by combining information from model runs and observational data. The existing standard calibration framework suffers from inferential issues when the model output and observational data are high-dimensional dependent data such as large time series due to the difficulty in building an emulator and the non-identifiability between effects from input parameters and data-model discrepancy. To overcome these challenges we propose a new calibration framework based on a deep neural network (DNN) with long-short term memory layers that directly emulates the inverse relationship between the model output and input parameters. Adopting the ‘learning with noise’ idea we train our DNN model to filter out the effects from data model discrepancy on input parameter inference. We also formulate a new way to construct interval predictions for DNN using quantile regression to quantify the uncertainty in input parameter estimates. Through a simulation study and real data application with WRF-hydro model we show that our approach can yield accurate point estimates and well calibrated interval estimates for input parameters.  
\end{abstract}

\noindent%

\spacingset{1.45} 

\section{Introduction}

Computer models play an important role in almost every field of science and engineering. These models are typically a collection of a large number of partial differential equations designed to capture the behavior of a real world process. These models typically have a set of uncertain input parameters that need to be properly calibrated using real data to generate realistic simulation. Since the seminal paper by \cite{kennedy2001bayesian} there has been a considerable growth in the literature of compute model calibration \citep[e.g.][]{bayarri2007computer,Higdon2008,wong2017frequentist,tuo2015efficient,chang2015binary,wong2017frequentist,salter2019uncertainty,sung2020generalized}.

The methodological challenges in this area can be summarized into two aspects. The first aspect stems from the fact that computer model runs are often available only at a limited number of design points. This leads to the need of a statistical surrogate (``emulator'')  for the computer model in question, typically done by constructing a Gaussian process (GP) model that interpolates computer model outputs at input parameter settings for which the model runs are not obtained \citep{sacks1989design}.
This issue is further complicated by the fact that modern computer model outputs are usually in the form of high-dimensional data with a complicated dependence structure such as large time series or spatial data. Building a GP emulator for such data poses considerable statistical and inferential challenges \citep{Higdon2008,chang2013fast,gu2016parallel,salter2019uncertainty} and the amount of effort to address these challenges often exceeds that to solve the calibration problem itself. 

The second aspect comes from the fact that most computer models are imperfect in representing the reality and hence one can reasonably expect that there is considerable discrepancy between the computer model output and the corresponding real world observation. When the model output is in the form of complicated dependent data such as time series the corresponding data-model discrepancy also likely has a complex dependent structure. If not handled properly this problem can cause significant bias in input parameter estimation. The existing methods rely on problem specific solutions such as assuming a prior distribution \citep{brynjarsdottir2014learning} or regularizing the complexity of the discrepancy term \citep{chang2013fast,tuo2015efficient}. However such solutions require substantial knowledge or specific assumption about the form of discrepancy, which are not always available or justifiable.

In this paper we propose an alternative framework to the existing calibration approach that takes an advantage of the recent development in deep neural network (DNN) methodologies. Our focus is on calibration using time series data, which are one of the most common form of computer model output \citep{bayarri2007computer,Higdon2008}, but the basic framework can be easily modified to other types of data such as spatial data. The main idea is to build a DNN model that can ``predict'' the optimal input parameter values for a given observational data by emulating the inverse relationship between the model output and input parameter values. To effectively filter out the effect of possible data-model discrepancy without imposing a strong assumption on the discrepancy term we adopt the idea of `learning with noise' \citep{koistinen1992kernel,holmstrom1992using,bishop1995training,an1996effects,vincent2010stacked}. In combination with the feature extraction capability of the modern DNN architecture this approach allows us to train a DNN model that can focus on the features that are relevant to parameter estimation while negating the effect of discrepancies. 

In addition to the new calibration framework we propose a new way to quantify uncertainty in prediction using DNN. Computer model calibration requires not only estimating the optimal values for the input parameters but also quantifying the surrounding uncertainties. Uncertainty quantification for DNN predictions is in general challenging because a DNN typically contains a large number of model parameters and it has been unclear how to reflect uncertainties in those parameters when constructing interval predictions without relying on some variational approximation to the likelihood function \citep{gal2016dropout}. We propose a quantile regression approach based on the observation that a DNN can be viewed as a linear regression with basis functions that are created by hidden layers. Our simulation study shows that this approach provides a better way to quantify the uncertainty as it is not prone to overconfidence issues that variational approximation-based approaches typically suffer from. To demonstrate that our method can efficiently estimate input parameters in a complicated modern computer model we apply our method to WRF-Hydro, a recently developed hydrologic module for the weather research and forecast (WRF) model \citep{gochis2015WRFHydro}.

The remainder of the paper is organized as follows: Section \ref{sec:KOHframework} describes the existing standard calibration framework and explains the common inferential challenges faced by the approach. Section \ref{sec:CalibrationUsingDNN} introduces our new inverse model-based framework using DNN that can overcome the challenges described in Section \ref{sec:KOHframework}. Section \ref{sec:inference} describes the details of inference procedure for our calibration method including regularized optimization and uncertainty quantification with quantile regression. Section 5 describes simulation study and Section 6 shows an example application of our approach to WRF-Hydro model. Section 7 summarizes the findings from our work and discusses future research directions.
  
\section{Standard Calibration Framework and its Challenges}
\label{sec:KOHframework}
We first define notation for the model output, input parameters and observational data to facilitate our discussion on the methodological development. Let $\bY(\btheta)$ be a $p$-dimensional model output at an input parameter setting $\btheta \in \mathcal{R}^{d_{\theta}}$. The output $\bY(\btheta)$ is typically in the form of spatial or temporal or spatio-temporal data. 
We also let $\bZ=[Z_1,\dots,Z_p]^T$ be the $p$-dimensional observational data that have the same format as the model output $\bY(\btheta)$. Throughout the rest of this paper we focus on the situation where both $\bY(\btheta)$ and $\bZ$ are temporal data. Since in most scientific applications obtaining the model output $\bY(\btheta)$ at each input parameter setting $\btheta$ is computationally expensive, model outputs are obtained at a limited number of design points $\btheta_1,\dots,\btheta_n$ with $n$ being typically hundreds or thousands. The resulting collection of model outputs $\bY(\btheta_1),\dots,\bY(\btheta_n)$ is often called a `perturbed physics ensemble.' 

The objective of statistical computer model calibration is to infer the realistic value for the input parameter $\btheta^*$ given the observational data $\bZ$ and the model outputs $\bY(\btheta_1),\dots,$ $\bY(\btheta_n)$. In other words, our objective is to find the best input parameter setting $\btheta^*$ for $\bZ$ given the observed relationship between $\btheta$ and $\bY(\btheta)$ from the perturbed physics ensemble. This problem therefore can be viewed as a classification problem with `continuous labels' $\btheta$. For our scientific problem described in Section \ref{sec:application} the number of model runs is 400 ($n=400$), the size of each model run and observational data is 480 ($p=480$), and the dimensionality of individual input parameter setting $d_\theta$ is 5 ($d_\theta=5$). 

\subsection{Existing Forward Model-based Approach}
\label{sec:ForwardMethod}
In this section we describe the existing standard computer model calibration framework that is currently widely used in the statistical literature. The standard computer model calibration model described in \cite{kennedy2001bayesian} can be written as
\begin{equation} \label{eqn:ForwardFramework}
    \bZ= \bY(\btheta^*)+\bdelta,
\end{equation}
where $\bdelta$ represents the data-model discrepancy often modeled by a $p$-dimensional Gaussian process (GP). The discrepancy includes both the structural error in the computer model (i.e. misrepresentation of the reality by the computer model) and the measurement error in observational data. The data type for $\bZ$ and $\bY(\btheta)$ determines the form of covariance function for $\bdelta$. Here we assume that $\bZ$ and $\bY(\btheta)$ are time series and hence the discrepancy term is also a time series that can be denoted as  $\bdelta=\left[\delta_1, \delta_2, \dots,\delta_t,\dots,\delta_p \right]$. In this case a 1-dimensional \matern ~class or an autoregessive model  can be used as a model for $\bdelta$. The likelihood function based on \eqref{eqn:ForwardFramework} can be used for inferring $\btheta^*$, while accounting for possible data-model discrepancies and observational errors. Evaluating the likelihood function based on the model in \eqref{eqn:ForwardFramework} requires running the forward model $\bY(\cdot)$ for the given value of $\btheta^*$ and hence we call this method a `forward model-based calibration'. If the forward model $\bY(\cdot)$ is computationally expensive and the evaluated model output $\bY(\btheta)$ is available at only a limited number of input parameter settings, which is the case for most scientific problems including the problem described in Section \ref{sec:application}, an emulator $\boldeta(\btheta)$ that approximates the forward model  $\bY(\btheta)$ is used instead. The emulator is typically constructed based on model runs $\bY(\btheta_1),\dots,\bY(\btheta_n)$ obtained at pre-specified design points $\btheta_1,\dots,\btheta_n$ using a GP model  \citep{sacks1989design}. 

\subsection{Challenges in Existing Framework}
\label{sec:challenges}
The forward model-based calibration framework described above often faces two important inferential and computational challenges: First, in most calibration problems we need to construct an emulator $\boldeta(\btheta)$ that can accurately predict the model output $\bY(\btheta)$ at any given new $\btheta$ that is not tried in the existing ensemble $\bY(\btheta_1),\dots,\bY(\btheta_n)$. This is often challenging especially when the model output $\bY(\btheta)$ exhibits a complicated dependence structure. Such problem is often further complicated by the usual `big' data issues for GP-based methods, i.e. the likelihood evaluation becomes computationally slow or even infeasible due to the difficulty in taking a cholesky decomposition of a large covariance matrix \citep{Higdon2008,chang2013fast,chang2013composite} when the model output is in the form of high-dimensional dependent data such as large time series. The computational complexity for each likelihood evaluation scales as $\mathcal{O}(p^3)$.

Second, the effects from the input parameter $\btheta^*$ and the effects from the data-model discrepancy $\bdelta$ cannot be identifiable in general and hence lead to biased or overly uncertain estimates for $\btheta^*$ \citep{brynjarsdottir2014learning, tuo2015efficient, salter2019uncertainty}. In particular, if the observational data appear to be quite different from any of the model runs due to data model discrepancy, parameter estimation results can be severely biased \citep[mentioned as `terminal case' in][]{salter2019uncertainty} as a zero-mean discrepancy term $\bdelta$ cannot easily capture such a trend.  
This also often leads to incorrect uncertainty quantification with poorly calibrated interval estimates for target input parameters, potentially resulting in a severe undercoverage of interval estimates. 


\section{Inverse Model-Based Calibration using DNN} \label{sec:CalibrationUsingDNN}
\subsection{Inverse Model-Based Calibration Framework}
\label{sec:InverseModelBasedCalibration}
In this section we propose our new inverse model-based calibration method using a deep neural network that can overcome the aforementioned challenges in the existing forward model-based calibration method. The main idea is to find the inverse function $\boldsymbol{g}$ that provides the best input parameter setting $\btheta^*$ when the observational data $\bZ$ is given, i.e.,
\begin{equation} \label{eqn:basic}
\btheta^*=\boldsymbol{g}(\bZ)+\bepsilon,
\end{equation}
with some $d$-dimensional prediction error term $\bepsilon$. Finding such function $\boldsymbol{g}$ can be thought as finding a function that satisfies
\begin{equation} \label{eqn:estimation}
\btheta=\boldsymbol{g}(\bY(\btheta)+\bdelta)+\bepsilon,
\end{equation}
for any $\btheta \in \Theta$ where $\Theta$ is the possible range for $\btheta^*$. In other words our objective is to find a function $\boldsymbol{g}$ that can filter out the discrepancy $\bdelta$ and accurately estimate $\btheta$ that originally generated  $\bY(\btheta)$ in any given observation $\bY(\btheta)+\bdelta$. Given the estimated function $\hat{\boldsymbol{g}}$ based on the model in \eqref{eqn:estimation}, the best predicted parameter setting $\btheta^*$ can be simply computed by
\begin{equation*} 
\hat{\btheta}^*=\hat{\boldsymbol{g}}(\bZ).
\end{equation*}

The approximation function $\hat{\boldsymbol{g}}$ has to possess the following properties: First of all, $\hat{\boldsymbol{g}}$ needs to be able to capture a highly nonlinear relationship, which is almost always expected in computer model calibration problems. In addition $\hat{\boldsymbol{g}}$ needs to be able to handle high-dimensional predictor variables with a complicated dependence structure such as long time series or large spatial data \citep[see, e.g.,][]{Higdon2008,chang2013fast,chang2015binary,bayarri2007computer,gu2016parallel,guan2019computer,sung2020generalized} because modern computer models commonly generate such type of data as their output. Another consideration is noise filtering: the function $\hat{\boldsymbol{g}}$ needs to be able to recover $\btheta$ from a noisy model output $\bY(\btheta)+\bdelta$ by filtering out the effects from the discrepancy $\bdelta$. 

In this paper we use a DNN to find the approximation function $\hat{\boldsymbol{g}}$. This choice is natural because DNN models possess all three required characteristics above. The main feature of DNN is its ability to approximate highly complicated non-linear functions, which has been proven in a wide range of applications and also discussed in some approximation theory point of view \citep[e.g.,][]{poggio2017and, chen2019efficient, schmidt2017nonparametric}. Moreover the recently developed architectures in DNN 
such as long-short term memory (LSTM) network \citep[e.g.,][]{huang2015bidirectional} can provide well-proven recipe for extracting important features from large time series data. The recently developed computational machineries including back-propagation and stochastic gradient descent algorithms facilitate easy implementation of DNN with a highly complicated structure. In the following subsections we explain the details of our DNN-based method for computer model calibration.

\subsection{DNN for Nonlinear Regression with Feature Extraction}
The most commonly used DNN architecture consists of two components: feature extraction layers and non-linear regression layers. The feature extraction layers apply a series of transformation to the input data set to find the `features' that are most relevant to predicting the response variables. For our calibration problem, the `features' found by the feature extraction layers can be interpreted as transformed data that are most relevant to estimating the input parameter setting. The non-linear regression layers create a non-linear function that links the extracted features to the response variables. In our calibration problem the non-linear regression layers estimate the best input parameter setting given the extracted features from data. 

The form of feature extraction layers is determined by the data type of the model output $\bY(\btheta)$ and the observational data $\bZ$. Since our focus here is on time series data   the suitable feature extraction model will be a bidirectional LSTM network \citep{huang2015bidirectional}. This structure combines information from the `forward' and `backward' LSTM units, where forward LSTM units model the information flow in time order and backward LSTM units model the information flow in reverse time order. This structure has been proven to be useful in capturing important features for sequence classification. 
The overall structure of the DNN structure described in this section is illustrated in Figure \ref{fig:bi-lstm}. (Note that sections, figures and tables referred with prefix S henceforth can be found in the Supplementary Document.)

One important advantage of this approach is computational complexity, which is scaled as $\mathcal{O}(p^2)$ \citep{sak2014long}. The difference in computing time between the DNN based method with LSTM and the GP-based method described in Section \ref{sec:ForwardMethod} grows exponentially as the size of model output $p$ grows, because the computational complexity of the GP-Fwd method scales as $\mathcal{O}(p^3)$ as discussed in Section \ref{sec:challenges}.

\subsubsection{Long-short Term Memory for Feature Extraction} \label{sec:LSTMlayers}
Recurrent neural networks (RNN) are neural networks specialized in handling sequential data. The hidden layers in Recurrent neural network (RNN) are connected in a cyclic pattern or self-connected loop. The LSTM (Hochreiter and Schmidhuber, 1997; Gers et al.,2000) network is currently the most widely used recurrent neural network for various applications including speech recognition, natural language processing, and sentiment analysis \citep[e.g.,][]{graves2005framewise,sak2014long,wang2016attention}. The main advantage of LSTM network is its ability to handle both short-range and long-range dependence in a computationally efficient manner. Moreover, the `gated' structure of LSTM that regulates the information flow within the network is helpful for avoiding computational issues (see Section \ref{sec:model_fitting} for further discussion) 

An LSTM takes a sequence as input and pass it through connected hidden layers to yield estimated values as output at each time point. To be more specific for a given $d_x$ dimensional input vector $\bx_t$ at each time point $t$ the $d_c$-dimensional `cell' vector $\overrightarrow{\bc}_t$ and its corresponding $d_c$-dimensional output vector $\overrightarrow{\mathbf{h}}_t$ are computed as
\begin{align}
\begin{split}
	\overrightarrow{\bc}_t=& \overrightarrow{\mathbf{u}}_{t}^{(f)} * \overrightarrow{\mathbf{c}}_{t-1} + \overrightarrow{\mathbf{u}}_{t}^{(i)} *  \mathbf{f}^{(c)}\left( \overrightarrow{\bW}^{(c)}_x \bx_t + \overrightarrow{\bW}^{(c)}_h \overrightarrow{\bh}_{t-1} + \overrightarrow{\mathbf{a}}^{(c)}\right), \label{eqn:LSTM} \\
\overrightarrow{\bh}_t=& \overrightarrow{\mathbf{u}}_{t}^{(o)} * \mathbf{f}^{(h)}(\overrightarrow{\bc}_t), 
\end{split}
\end{align}
where $\overrightarrow{\bW}_x^{(c)}$ and $\overrightarrow{\bW}_h^{(c)}$ are respectively $d_c \times d_x$ and $d_c \times d_c$ weight matrices for input $\bx_t$ and output from previous time step $\overrightarrow{\bh}_{t-1}$; $\overrightarrow{\mathbf{a}}^{(c)}$ is a $d_c$-dimensional intercept vector (often called `bias' in the deep learning literature); $\mathbf{f}^{(c)}| R^{d_c}\rightarrow R^{d_c}$ and $\mathbf{f}^{(h)}|R^{d_c} \rightarrow R^{d_c}$ are `activation' functions for nonlinear transformation. Here the arrow $\overrightarrow{\cdot}$ is used to emphasize that the matrices and vectors are for a network that models information flow going forward in time. (Below a network for backward flow will be introduced as well.) The initial values $\overrightarrow{\bc_0}$ and $\overrightarrow{\bh_0}$ are set to be zeros.  The operator $*$ denotes element-wise multiplication and $\overrightarrow{\bu}_t^{(f)}$, $\overrightarrow{\bu}_t^{(i)}$, and $\overrightarrow{\bu}_t^{(o)}$ are respectively the \emph{forget}, \emph{input} and \emph{output} `gate' vectors (thereafter shortened as `gate'). The gates are defined in a similar fashion as a usual neural network node:
\begin{align}
\begin{split} \label{eqn:gates}
\overrightarrow{\bu}_t^{(f)}=& \mathbf{f}^{(f)}\left(\overrightarrow{\bW}_x^{(f)} \bx_t + \overrightarrow{\bW}_h^{(f)} \overrightarrow{\bh}_{t-1} +\overrightarrow{\mathbf{a}}^{(f)}\right),\\
\overrightarrow{\bu}_t^{(i)}=& \mathbf{f}^{(i)}\left(\overrightarrow{\bW}_x^{(i)} \bx_t + \overrightarrow{\bW}_h^{(i)} \overrightarrow{\bh}_{t-1} +\overrightarrow{\mathbf{a}}^{(i)}\right), \\
\overrightarrow{\bu}_t^{(o)}=& \mathbf{f}^{(o)}\left(\overrightarrow{\bW}_x^{(o)} \bx_t + \overrightarrow{\bW}_h^{(o)} \overrightarrow{\bh}_{t-1} +\overrightarrow{\mathbf{a}}^{(o)}\right),
\end{split}
\end{align}
where matrices denoted as $\overrightarrow{\bW}_x^{(.)}$ and $\overrightarrow{\bW}_h^{(.)}$ are respectively $d_c \times d_x$ and $d_c \times d_c$ weight matrices that link input variables $\bx_t$ and previous output $\overrightarrow{\bh}_{t-1}$ to each gate vector; vectors denoted as  $\overrightarrow{\mathbf{a}}^{(.)}$ are $d_c$-dimensional intercept vectors for each gate; functions denoted as $\mathbf{f}^{(.)}| R^{d_c}\rightarrow R^{d_c}$ are activation functions for each gate. These gates control how the information flows within the LSTM network and including them improves numerical stability as well as prediction accuracy \citep{gers2001lstm}.  The input vector $\bx_t$ is defined as the current and lagged variables of observed sequence, i.e. $\bx_t=[Z_{t-{d_t}},\dots,Z_{t}]^T$, which supplies information from short range time dependence (or `short term memory') to the network. The sequential cells  $\overrightarrow{\bc}_1,\dots,\overrightarrow{\bc}_p$ are designed to capture the long range dependence (or `long term memory') in the modeled time sequence. 

If our goal was to make predictions on the observed sequence $Z_t$, the models described in \eqref{eqn:LSTM} and \eqref{eqn:gates} would be enough. However, since our goal here is to extract features from the observed sequence and use it for finding the best value for $\btheta$, a bidirectional LSTM network is more suitable \citep{huang2015bidirectional}. In addition to the forward LSTM layers described in \eqref{eqn:LSTM} and \eqref{eqn:gates} we have the following backward LSTM layers at each time step $t$:
\begin{align}
\begin{split} 
	\overleftarrow{\bc}_t=& \overleftarrow{\mathbf{u}}_{t}^{(f)} * \overleftarrow{\mathbf{c}}_{t+1} + \overleftarrow{\mathbf{u}}_{t}^{(i)} *  \mathbf{f}^{(c)}\left( \overleftarrow{\bW}_x^{(c)} \bx_t + \overleftarrow{\bW}_h^{(c)} \overleftarrow{\bh}_{t+1} + \overleftarrow{\mathbf{a}}^{(c)}\right), \label{eqn:LSTMback} \\
\overleftarrow{\bh}_t=& \overleftarrow{\mathbf{u}}_{t}^{(o)} * \mathbf{f}^{(h)}(\overleftarrow{\bc}_t), 
\end{split}
\end{align}
with the following gate structure that has the same form as the forward LSTM units:
\begin{align}
\begin{split}  \label{eqn:gatesback} 
\overleftarrow{\bu}_t^{(f)}=& \mathbf{f}^{(f)}\left(\overleftarrow{\bW}_x^{(f)} \bx_t + \overleftarrow{\bW}_h^{(f)} \overleftarrow{\bh}_{t+1} +\overleftarrow{\mathbf{a}}^{(f)}\right),\\
\overleftarrow{\bu}_t^{(i)}=& \mathbf{f}^{(i)}\left(\overleftarrow{\bW}_x^{(i)} \bx_t + \overleftarrow{\bW}_h^{(i)} \overleftarrow{\bh}_{t+1} +\overleftarrow{\mathbf{a}}^{(i)}\right), \\
\overleftarrow{\bu}_t^{(o)}=& \mathbf{f}^{(o)}\left(\overleftarrow{\bW}_x^{(o)} \bx_t + \overleftarrow{\bW}_h^{(o)} \overleftarrow{\bh}_{t+1} +\overleftarrow{\mathbf{a}}^{(o)}\right),
\end{split}
\end{align}
where the vectors and matrices in \eqref{eqn:LSTMback} and \eqref{eqn:gatesback} with the backward arrow $\overleftarrow{\cdot}$ have the same dimensionalities as their counterparts in \eqref{eqn:LSTM} and \eqref{eqn:gates} with the forward arrow $\overrightarrow{\cdot}$. Again, the initial values $\overrightarrow{\bc_{T+1}}$ and $\overrightarrow{\bh_{T+1}}$ are set to be zeroes. The output vectors from the backward and forward LSTM units at each time step $t$ are summed into one output vector $\bh_t$,
\begin{equation}
    \bh_t=\overrightarrow{\bh}_{t}+\overleftarrow{\bh}_{t},
\end{equation}
which will be passed to the non-linear regression layers. Figure \ref{fig:bi-lstm} illustrates the resulting LSTM structure. 

The activation functions $\mathbf{f}^{(\cdot)}$ are defined as a collection of 1-dimensional functions $f_i^{(\cdot)}|R\rightarrow R$ $(i=1,\dots,d_c)$ as follows:
\begin{equation*}
\mathbf{f}^{(\cdot)}(\cdot)=\left[f_1^{(\cdot)}(\cdot),f_2^{(\cdot)}(\cdot),\dots,f_{d_{c}}^{(\cdot)}(\cdot)\right]^T.
\end{equation*}
Following a typical choice in the literature we use the `hard sigmoid' function for the activation functions $\mathbf{f}^{(f)}$, $\mathbf{f}^{(i)}$, $\mathbf{f}^{(o)}$ for the gate variables, i.e.
\begin{equation*}
f_i^{(.)}(x)=\max(0,\min(1,x))
\end{equation*}
for $f_i^{(f)}$, $f_i^{(i)}$, and $f_i^{(o)}$ ($i=1,\dots,d_c$). This choice sets a large number of values in the gate vectors in both forward and backword LSTM layers ($\overrightarrow{u}_t^{(f)}$, $\overrightarrow{u}_t^{(i)}$, $\overrightarrow{u}_t^{(o)}$, $\overleftarrow{u}_t^{(f)}$, $\overleftarrow{u}_t^{(i)}$ and $\overleftarrow{u}_t^{(o)}$) to be zeros, and hence imposes a strong  regularization through sparsity. For the remaining activation functions $\mathbf{f}^{(c)}$ and $\mathbf{f}^{(h)}$ we use the rectified linear unit \citep[ReLU, see e.g.][Ch. 6]{goodfellow2016deep}:
\begin{equation} \label{eqn:ReLU}
f_i^{(.)}(x) = \max(0,x)
\end{equation}
for individual $f_i^{(c)}$ and $f_i^{(h)}$ ($i=1,\dots,J_c$). It is well known that using ReLU as activation functions greatly increases numerical stability in likelihood estimation for deep neural networks. We discuss the rationale behind this choice in detail in Section \ref{sec:model_fitting}.

\subsubsection{Fully-Connected Layers for Nonlinear Regression} \label{sec:FClayers}

The final outputs from the feature extraction layers are vectorized (often referred to as `flattening' in the deep learning literature) as $\blambda^{(0)}=[\bh_{1}^T,\dots,\bh_{p}^T]^T$ and supplied to the nonlinear regression layers. We use a fully connected network with $L$ layers as our nonlinear regression layers. The model structure for the fully connected layers can be written as
\begin{align} 
\begin{split}
\label{eqn:fcl}
\blambda^{(1)} &=\mathbf{f}^{(1)}\left(\bW^{(0)} \blambda^{(0)} +\mathbf{a}^{(0)}\right),\\
\blambda^{(2)} &=\mathbf{f}^{(2)}\left(\bW^{(1)} \blambda^{(1)}+\mathbf{a}^{(1)}\right),\\
&\dots\\
\blambda^{(L)} &=\mathbf{f}^{(L)}\left(\bW^{(L-1)} \blambda^{(L-1)}+\mathbf{a}^{(L-1)}\right),\\
\hat{\btheta}^* &=\bW^{(L)} \blambda^{(L)}+\mathbf{a}^{(L)},
\end{split}
\end{align}
where $\blambda^{(l)}$ is the vector for the $d_{(l)}$ different nodes in the $l$th layer;  $\mathbf{f}^{(l)}| R^{d_{(l)}} \rightarrow R^{d_{(l)}}$ is a vector-valued activation function for the $l$th layer; $\bW^{(l)}$ is a $d_{(l+1)}\times d_{(l)}$ weight matrix; $\ba^{(l)}$ is a $d_{(l+1)}$-dimensional intercept matrix (which is often called `bias' in the deep learning literature). The length of $\blambda^{(0)}$ (i.e., $d_{(0)}$) is determined as $T d_c$ because the length of each $\bh_t$ is $d_c$. The sizes of subsequent layers, $d_{(1)},\dots,d_{(L)}$, which are often referred to as the widths of layers, need to be determined by the user.  The width of the last layer $d_{(L+1)}$ is $d_\theta$ (the dimensionality of $\hat{\btheta}^*$) and hence $\bW^{(L)}$ is a $d_\theta \times d_{(L)} $ matrix and $\ba^{(L)}$ is a $d_\theta$-dimensional vector. 

The recent development in approximation theories \citep[e.g.,][]{poggio2017and, chen2019efficient, schmidt2017nonparametric} suggest that having multiple hidden layers (i.e., $L\gg 1$) to build a `deep' network leads to a better prediction performance for the response variable than having a shallow network, coining the term `deep learning'. Having a deep network however poses a danger of `saturation' or `vanishing gradient', meaning that the gradient of the resulting likelihood function becomes zero for a wide range of predictor variables and hence gradient-based optimization methods such as gradient descent search become computationally infeasible. (see Section \ref{sec:MLE} below for further discussion). This issue can be avoided by choosing a proper activation function: for the $l$th layer activation function $\mathbf{f}^{(l)}(\cdot)=\left[f_1^{(l)}(\cdot),f_2^{(l)}(\cdot),\dots,f_{d_{(l)}}^{(l)}(\cdot)\right]^T$ we define the activation function as ReLU defined in \eqref{eqn:ReLU}. This choice of activation function also imposes certain level of `sparsity' to the network by making a large portion of $\blambda^{(l)}$ become zeros.

\subsection{Handling Data-Model Discrepancy}
\label{sec:discrepancy}
In our calibration approach the main goal of statistical inference is to build an inverse function $\hat{\boldsymbol{g}}$ that can efficiently estimate $\btheta$ from $\bY(\btheta)+\bdelta$ even under the presence of data-model discrepancy $\bdelta$. This problem resembles the problem of noisy sequence classification except that the response variable is a continuous variable in our case. Inspired by the idea of `learning with noise' in the neural network literature \citep{koistinen1992kernel,holmstrom1992using,bishop1995training,an1996effects,vincent2010stacked} we propose to train the inverse emulator $\hat{\boldsymbol{g}}$ using `contaminated' model outputs instead of the original model outputs. In this way the resulting neural network model $\hat{\boldsymbol{g}}$ can automatically extract the features $\blambda^{(0)}$ from a noisy model output $\bY(\btheta)+\bdelta$ that is most relevant to recovering the input parameter setting $\btheta$.  

To this end we generate $n_d$ different realizations of $\bdelta$ from an assumed discrepancy distribution for each input parameter setting $\btheta_i$ $(i=1,\dots,n)$ to have generated discrepancy terms $\left\{ \bdelta_{ij} \right\}$ ($i=1,\dots,n$ and $j=1,\dots,n_d$). We then create contaminated model outputs $\tilde{\bY}_1,\dots,\tilde{\bY}_N$ with $N=n\times n_d$ by superimposing the generated discrepancy terms on the original model outputs as follows: 
\begin{equation*}
    \tilde{\bY}_k=\bY(\btheta_{i})+\bdelta_{ij}
\end{equation*}
for $i=1,\dots,n$ and $j=1,\dots,n_d$, where $k=n_d(i-1)+j$. We let $\tilde{\btheta}_1,\dots,\tilde{\btheta}_N$ denote the input parameter settings used for creating $\tilde{\bY}_1,\dots,\tilde{\bY}_N$ (i.e., $\tilde{\btheta}_k = \btheta_{\lceil k/n_d \rceil}$). This `learning with error' approach aims to train the DNN model with various types of data-model discrepancy patterns so that it can handle discrepancies varying in a wide range of magnitudes and time scales.

For the discrepancy model for $\bdelta$ we use a zero mean Gaussian process model with the following squared exponential covariance function for generating $\bdelta_{ij}$: 
\begin{equation*}
    Cov(\delta_{t_1},\delta_{t_2})= \zeta  1(t_1=t_2) + \kappa \exp\left(-\frac{\left|\delta_{t_1}-\delta_{t_2}\right|^2}{\phi}\right),
\end{equation*}
where $t_1, t_2 \in\left\{1,\dots,T\right\}$, $1(\cdot)$ is an indicator function for the condition in $(\cdot)$; $\zeta>0$, $\kappa>0$, and $\phi>0$ are respectively the nugget, partial sill, and the range parameters. To avoid imposing a too strong assumption on the discrepancy term we allow these parameters to vary across different realizations of $\bdelta_{ij}$ so that the resulting inverse function $\hat{g}$ can handle various types of $\bdelta$ patterns. We generate a sample of size $N$ for these parameter values based on a Latin hypercube design. Ranges for the parameters (preferably broad) are the only required input. The ranges for $\zeta$ and $\kappa$ reflect model user's guess on the magnitudes of independent and time-dependent components in the data-model discrepancy. (See Section \ref{sec:ImplementationDetails} for the specific parameter ranges used in our application problem.)  As per the range of $\phi$, one rule that can be used for a wide range of problems is to use a value between 1\% and 10\% of the time interval lengths ($p$) as the lower limit and a value between 60\% and 70\% of the length as the upper limit so that the generated discrepancy patterns cover various types of structured errors including errors with short range dependence (when $\phi$ is near its lower limit) and overall mean shift (when $\phi$ is near its upper limit). One can choose a more informative sampling scheme that puts more emphasis on certain parts of the discrepancy parameter space if some prior knowledge that justifies such choice exists for the problem at hand. 

\section{Statistical Inference for DNN Calibration}
\label{sec:inference}
We now describe the details of inference for the calibration model proposed in Section \ref{sec:CalibrationUsingDNN}. We illustrate how the model is fitted with a proper regularization and how the input parameters are predicted along with their uncertainty intervals. 

\subsection{Minimizing the Stochastic Loss Function with Dropout} \label{sec:MLE}

In this section we use $\hat{\boldsymbol{g}}(\cdot)$ to exclusively denote the approximation function constructed by the deep network explained in Sections \ref{sec:LSTMlayers} and \ref{sec:FClayers}. We also let $\bw=\left[w_1,\dots,w_{n_w}\right]^T$ denote a vector of all parameters contained in weight matrices and intercept vectors defined in \eqref{eqn:LSTM}, \eqref{eqn:gates}, \eqref{eqn:LSTMback}, \eqref{eqn:gatesback}, and \eqref{eqn:fcl} where $n_w$ is the total number of parameters in the deep network model. The inference problem here is to estimate quantities in $\bw$ based on $\tilde{\bY}_1,\dots,\tilde{\bY}_N$. The standard `cost' function used in the deep learning literature for continuous response variables is the \emph{squared loss function} given as
\begin{equation*} 
    \mathcal{L}(\bw)=\sum_{i=1}^N (\hat{\btheta}_i^*-\tilde{\btheta}_i)^T(\hat{\btheta}_i^*-\tilde{\btheta}_i),
\end{equation*}
where $\hat{\btheta}_i^*=\hat{\boldsymbol{g}}(\tilde{\bY}_i)$. 
Minimizing this cost function is equivalent to maximizing the log-likelihood function for the model in \eqref{eqn:basic} with an assumption $\bepsilon\sim N(\mathbf{0},\sigma^2 \mathbf{I}_{d_\theta})$ with $\sigma^2>0$ (i.e., assuming equal variance for $\bepsilon$). Here the equal variance assumption for the $d_\theta$ different input parameters  can be justified by rescaling the input parameters so that they have the same range (typically [0,1]) and hence operate in the same scale. 

The deep network model described in Sections \ref{sec:LSTMlayers} and \ref{sec:FClayers} is apparently over-parametrized and it is often helpful to impose some regularization for a better prediction performance \citep[][Chapter 7]{goodfellow2016deep}. We implement two approaches simultaneously that are frequently used in the deep learning literature: dropout and penalized likelihood. 

Dropout is a way to create a stochastic likelihood function by introducing some randomness in the structure of our deep network \citep[][Chapter 7.12]{goodfellow2016deep}. To be more specific we re-define the fully connected layers in \eqref{eqn:fcl} as
\begin{align}
\begin{split} \label{eqn:fclDropout}
\blambda^{(1)} &=\mathbf{f}^{(1)}\left(\bW^{(0)}\blambda^{(0)}  +\mathbf{a}^{(0)}\right),\\
\blambda^{(2)} &=\mathbf{f}^{(2)}\left(\bW^{(1)} \blambda^{(1)}*\br^{(1)}+\mathbf{a}^{(1)}\right),\\
&\dots\\
\blambda^{(L)} &=\mathbf{f}^{(L)}\left(\bW^{(L-1)} \blambda^{(L-1)}*\br^{(L-1)}+\mathbf{a}^{(L-1)} \right),\\
\hat{\btheta}^* &=\bW^{(L)}\blambda^{(L)}*\br^{(L)} +\mathbf{a}^{(L)},
\end{split}
\end{align}
where $\br^{(l)}$ for $l=1,2,\dots,L$ is defined as  $d_{(l)}$-dimensional vectors whose elements are identically and independently distributed Bernoulli random variables with a pre-specified success probability $p_{keep}$. We let $\br$ denote a collection of all $\br^{(l)}$'s, i.e., $\br=\left[{\br^{(1)}}^T,{\br^{(2)}}^T,\dots,\right.$ $\left.{\br^{(L)}}^T\right]^T$. The operator $*$ denotes element-wise multiplication. This leads to a stochastic loss function since new values of $\br$ are drawn for every evaluation of the function. The loss function with dropout  $\mathcal{L}_{\br}(\bw)$ can be redefined as 
\begin{equation} \label{eqn:loss_function}
   \mathcal{L}_\br(\bw)=\sum_{i=1}^N (\hat{\btheta}_{\br,i}^*-\tilde{\btheta}_{i})^T(\hat{\btheta}_{\br,i}^*-\tilde{\btheta}_{i}). 
\end{equation}
where $\hat{\btheta}_{\br,i}^*=\hat{\boldsymbol{g}}_\br\left(\tilde{\bY}_i\right)$ and $\hat{\boldsymbol{g}}_\br$ is the deep learning-based approximation function constructed based on \eqref{eqn:fclDropout} instead of \eqref{eqn:fcl}. The subscript $\br$ is used to emphasize the dependence of the predicted values on the random vector $\br$. 

For penalization we can choose any commonly used form including lasso, ridge, and elastic net as the penalty function, which we will denote as $\mathcal{P}(\bw)$ henceforth. 
The resulting penalized loss function is given as
\begin{equation} \label{eqn:likelihood}
    \ell_{\br}(\bw) \propto \mathcal{L}_{\br}(\bw) + \mathcal{P}(\bw).
\end{equation}
One notable choice for $\mathcal{P}(\bw)$ in the literature \citep{gal2016dropout} is a ridge penalty term defined in \eqref{eqn:ridge} in Section \ref{sec:ridge_penalty}. 
When combined with dropout the resulting penalized likelihood function (i.e., the negative penalized loss function -$\ell_{\br}(\bw)$) can be thought as a variational approximation to the posterior of the deep Gaussian process model corresponding to our DNN model. (See Section \ref{sec:mc_dropout} and \cite{gal2016dropout} for further details.)


Note that dropout is applied only for parameter estimation, not prediction. In other words, once the parameter $\hat{\bw}$ is estimated by minimizing the loss function in \eqref{eqn:likelihood} the predictor $\hat{\btheta}^*$ is computed by the original model in \eqref{eqn:fcl} not \eqref{eqn:fclDropout}. An exception for this rule is when the MC dropout approach is applied \citep[][see Section \ref{sec:mc_dropout} for details]{gal2016dropout}. Model fitting using the penalized loss function in \eqref{eqn:likelihood} can be done through a gradient descent algorithm, which is described in Section \ref{sec:model_fitting}.

\subsection{Uncertainty Quantification Using Quantile Regression}
\label{sec:quantile}
The formulation in \eqref{eqn:basic} suggests that uncertainty quantification for the estimated input parameter $\btheta^*$ can be essentially boiled down to the problem of finding the prediction interval for the fitted DNN emulator $\hat{\boldsymbol{g}}$. However, the highly complicated structure of $\hat{\boldsymbol{g}}$ and a large number of parameters in $\bw$ make classic approaches to finding a prediction interval for $\btheta^*$ computationally prohibitive. For example, the asymptotic variance based on information matrix \citep{white1989some} cannot be computed because it requires inverting an $n_w \times n_w$ matrix and the total number of parameters $n_w$ is typically hundreds of thousands or more. Similarly, a fully Bayesian inference \citep[as mentioned in][]{polson2017deep} is not applicable either because it is not possible to fully explore the $n_w$-dimensional parameter space using Markov Chain Monte Carlo (MCMC).

To overcome the computational limitation we propose a quantile regression-based approach. Quantile regression has been used to construct prediction intervals for highly complex prediction models such as the random forest \citep[e.g.,][]{meinshausen2006quantile,zhang2019random}. The last equation in \eqref{eqn:fcl} suggests that the last layer of our DNN model can be viewed as a linear mean regression model between the response variable $\btheta^*$ and the extracted feature $\blambda^{(L)}$ up to the $L$th layer and consequently the predicted mean of $\btheta^*$ is
\begin{equation*}\label{eqn:mean}
\hat{\btheta}^* =\bW^{(L)} \blambda^{(L)}+\mathbf{a}^{(L)}.
\end{equation*}
A similar observation on DNN as a linear model with basis functions can be also found in \cite{mcdermott2019deep} and \cite{wikle2019comparison}. Instead, a predicted $\tau$th quantile of $\btheta^*$, denoted by  $\hat{\btheta}^*_\tau$, can be obtained by quantile regression:
\begin{equation*}\label{eqn:quantile}
\hat{\btheta}^*_\tau =\bW^{(L)}_\tau \blambda^{(L)}+\mathbf{a}^{(L)}_\tau,
\end{equation*}
where $\bW^{(L)}_\tau$ and $\mathbf{a}^{(L)}_\tau$ are the regression quantiles for a pre-specified target quantile $0<\tau<1$. The prediction limits are given as lower and upper tail quantiles such as the 0.025th ($\tau=0.025$) and the 0.975th ($\tau=0.975$) quantiles, that is  $\left[\hat{\btheta}^*_{0.025},\hat{\btheta}^*_{0.975}\right]$. Since the overall sample size $N$ is typically thousands or larger (see Sections  \ref{sec:simulation_study} and \ref{sec:application} below), these tail quantiles are reliably estimable. As noted at the end of Section \ref{sec:MLE} the upper and lower limits are computed without applying dropout. In addition to the interval estimates we can also find the median estimate $\hat{\btheta}^*_{0.5}$, a more robust estimate for $\btheta^*$ than the mean and use it as the point prediction.

We formulate the cost function to optimize based on \eqref{eqn:fclDropout} instead of \eqref{eqn:fcl} to apply dropout to this procedure.  
We propose to estimate the $\tau$th regression quantiles $\bW^{(L)}_\tau$ and $\mathbf{a}^{(L)}_\tau$ by conducting quantile regression only in the last layer of the DNN model in \eqref{eqn:fclDropout},
\begin{equation} \label{eqn:quantile_dropout}
    \hat{\btheta}_\tau^* =\bW_\tau^{(L)} \blambda^{(L)}*\br^{(L)}+\mathbf{a}_\tau^{(L)}
\end{equation}
while fixing the rest of the estimated parameters in the model at their optimal values obtained by minimizing the cost function \eqref{eqn:loss_function}. Therefore, the proposed quantile regression-based approach can be viewed as adding one more step in the end after completing the analysis in Section \ref{sec:MLE} to construct a prediction interval of $\btheta^*$.
(See Section \ref{sec:quantile_alternative} for rationale behind this.) Following the standard quantile regression procedure, we obtain the $\tau$th regression quantile estimate for $\bW^{(L)}_\tau$ and $\mathbf{a}^{(L)}_\tau$  by minimizing the following cost function:
\begin{equation} \label{eqn:cost_function_qunatile}
   \mathcal{L}_\br(\bW^{(L)}_\tau,\mathbf{a}^{(L)}_\tau)=\sum_{i=1}^N \left(\hat{\btheta}_{\tau,\br,i}^*-\tilde{\btheta}_{i}\right)^T\left[\tau \mathbf{1}_{d_\theta}-\boldsymbol{I}\left(\hat{\btheta}_{\tau,\br,i}^*-\tilde{\btheta}_{i}<\mathbf{0}\right)\right]
\end{equation} 
where $\mathbf{1}_{d_\theta}$ is a $d_\theta$-dimensional vector of 1's and $\boldsymbol{I}\left(\hat{\btheta}_{\tau,\br,i}^*-\tilde{\btheta}_{i}<\mathbf{0}\right)$ is a multivariate indicator function whose $j$th element is 1 if the $j$th element of $\hat{\btheta}_{\tau,\br,i}^*-\tilde{\btheta}_{i}$ is less than 0 or 0 otherwise for $j=1,\dots,d_\theta$. The estimated quantile $\hat{\btheta}_{\tau,\br,i}^*$ is defined as
\begin{equation*}
    \hat{\btheta}_{\tau,\br,i}^*=\hat{\boldsymbol{g}}_{\tau,\br}\left(\tilde{\bY}_i\right)
\end{equation*}
and $\hat{\boldsymbol{g}}_{\tau,\br}$ is the deep learning-based approximation function constructed by replacing the mean regression $\hat{\btheta}^* =\bW^{(L)}\blambda^{(L)}*\br^{(L)} +\mathbf{a}^{(L)}$ 
with the quantile regression  \eqref{eqn:quantile_dropout}. 





\section{Simulation Study} \label{sec:simulation_study}

In this section we verify the performance of our proposed DNN and quantile regression-based method (DNN-Q henceforth) and compare it with three other approaches through a simulation study using a synthetic computer model output and observational data. 

The first method to be compared is a DNN-based method that shares the framework introduced in Sections \ref{sec:CalibrationUsingDNN} and \ref{sec:inference} except for the uncertainty quantification method described in Section \ref{sec:quantile}. For uncertainty quantification this method employs MC dropout, an existing standard uncertainty quantification method for DNN \citep{gal2016dropout} based on variational Bayes approximation. We call this method DNN-MC henceforth. (See Section \ref{sec:mc_dropout} for details.) 

The second method to be compared is an inverse model-based approach that shares the same framework in Section \ref{sec:InverseModelBasedCalibration} but finds the estimated inverse function $\hat{\boldsymbol{g}}(\cdot)$ using the random forest. The random forest-based calibration approach has not been introduced in the literature before, but we compare our method to this approach to demonstrate that DNN provides a better  way to build $\hat{\boldsymbol{g}}(\cdot)$ than the random forest, which is also widely used as a general purpose function approximator. We call this method RF-Inv henceforth. (See Section \ref{sec:RFinv} for details.) 

The third method to be compared is the standard forward model-based calibration method explained in Section \ref{sec:ForwardMethod}. The method employs a Gaussian process emulator \citep{sacks1989design} to approximate the forward model $\bY(\btheta)$ and Bayesian inference to infer the best parameters $\btheta^*$ \citep{kennedy2001bayesian}. We call this method `GP-Fwd' for the rest of the manuscript. (See Section \ref{sec:GPfwd} for details.)

\subsection{Synthetic Model Outputs and Observational Data}
\label{SyntheticModelAndData}

By following the usual way of conducting simulation studies in the calibration method literature \citep[see., e.g.,][]{Higdon2008,chang2015binary} we generate synthetic model runs and observational data and try to learn the true input parameter settings for the synthetic observations using the synthetic model runs. 
To this end we first train the statistical emulators (either for the forward or inverse relationship) using the synthetic model runs and apply it to recover the parameter values for the synthetic observations. We compare the performance of all four methods in recovering the true input parameter settings for synthetic observations. 

We generate synthetic model outputs that have similar characteristics as the model outputs in Section \ref{sec:application}. For $p=480$ time points $t=1,\dots,480$ the model output $Y(\btheta,t)$ is defined as follows:
\begin{equation*}
    Y(\btheta,t)=0.3+\frac{\theta_1+0.3}{\sqrt{2\pi(\theta_3+0.1)}}\exp\left[-\frac{(u_t-\theta_2+0.5)^2}{\theta_3+0.1}\right]
\end{equation*}
where $u_1,\dots,u_{480}$ are equally spaced points starting from -2 to 2, $\btheta=[\theta_1,\theta_2,\theta_3]^T$ is a vector of the input parameters that governs how the synthetic model output behaves. The whole model output  at a given input parameter setting $\btheta$ can be denoted as $\bY(\btheta)=[Y(\btheta,1),\dots,Y(\btheta,480)]^T$. As shown in Figure \ref{fig:simul}, the synthetic model outputs are smooth curves on the interval $[0,480]$ with a single peak. The first parameter $\theta_1$ controls the overall scale of the output, the second parameter $\theta_2$ controls the location of the peak, and the third parameter $\theta_3$ controls the overall dispersion of the the curve. Based on this model we generate $n=200$ synthetic model runs that are used to build the inverse emulator $\hat{\boldsymbol{g}}(\cdot)$ in the DNN-Q, DNN-MC, and RF-Inv methods or the forward emulator $\boldeta(\cdot)$ in GP-Fwd method.
We also generate 1,500 different scenarios for synthetic observations $\bZ$, which serve as test data for model performance evaluation in Section \ref{sec:Results} below. (See Section \ref{sec:DataGenerationDetails} for details)

\subsection{Results}
\label{sec:Results}

We implement the four compared methods DNN-Q, DNN-MC, RF-Inv, and GP-Fwd based on the generated synthetic model runs and observational data. The implementation details are described in \ref{sec:ImplementationDetails}. We use four different metrics to compare the performance of different methods: the bias, the root mean square error (RMSE), the average length and the empirical coverage of the 95\% interval estimates for $\btheta^*$. The comparison results for the four methods are summarized in Tables \ref{tab:simul1}  and \ref{tab:simul2}. 

In Table \ref{tab:simul1} we compare the performance of the three inverse model-based approaches, DNN-Q, DNN-MC, and RF-Inv for the 1,500 test cases. The results show that all three methods provide decent point predictions with small biases and RMSEs for the test data set. Both DNN-based approaches have comparable RMSEs, lower than that of RF-Inv for all three parameters. In terms of uncertainty quantification through interval estimates DNN-Q yields empirical coverages that are close to the nominal confidence level for all three parameters for the test cases. DNN-MC yields much shorter prediction intervals than the other two methods but  has notable undercoverage issues for the first (0.803) and second parameters (0.893), which is often expected for a variational Bayes approximation. RF-Inv method shows notable undercoverage (0.823) for the first parameter. Moreover for all three parameters RF-Inv method leads to much wider average interval lengths for all three parameters compared to DNN-Q. Overall DNN-Q shows the most stable performance  without showing any notable undercoverages and with notably shorter prediction intervals than RF-Inv. 

\begin{table}[h]
    \caption{Simulation Study Results for all Test Cases}
    \begin{center}
    \begin{tabular}{c|c|c|c|c|c}
  Parameter &     Method & Bias & RMSE &  PI Length$^\dagger$  & PI Coverage$^\ddag$ \\
    \hline
  \multirow{3}{*}{$\theta_1$} &  DNN-Q & -0.002 & 0.056 & 0.204 &  0.923 \\
  &  DNN-MC &  0.000 & 0.053 & 0.116  & 0.803\\
  &  RF-Inv & 0.018  & 0.094 & 0.260  & 0.823 \\

    \hline
      \multirow{3}{*}{$\theta_2$}   & DNN-Q & -0.005 & 0.043 & 0.131  & 0.933 \\
  &  DNN-MC & -0.002 & 0.042 & 0.106& 0.893\\
  &  RF-Inv & -0.007 & 0.051 & 0.218 & 0.948\\
    \hline

      \multirow{3}{*}{$\theta_3$}   & DNN-Q & -0.001 & 0.034&  0.106& 0.931 \\
  &  DNN-MC & 0.002 & 0.030 & 0.108 & 0.954 \\
  &  RF-Inv & -0.001 &0.048  &0.194  & 0.957 \\
    \hline

    \end{tabular}
    \label{tab:simul1}
    \end{center}
    
    $\dagger$: Average Length of 95\% Prediction Interval.\\
    $\ddag$: Empirical Coverage of 95\% Prediction Interval.
\end{table}

\begin{table}[h]
    \begin{center}
    \caption{Simulation Study Results for Selected Cases for GP-Fwd}

    \begin{tabular}{c|c|c|c|c|c}
        \hline
    Data Set    & Method & Bias & RMSE &  PI Length$^\dagger$  & PI Coverage$^\ddag$ \\
    \hline
 \multirow{2}{*}{$\theta_1$} & DNN-Q & 0.005 & 0.041 & 0.208 & 0.980\\
   & GP-Fwd & -0.013 & 0.248 & 0.936 & 1.000 \\
    \hline
     \multirow{2}{*}{$\theta_2$} & DNN-Q & 0.002 & 0.037 & 0.128 & 0.940\\
   & GP-Fwd & -0.079 & 0.205& 0.071 & 0.280 \\
    \hline
     \multirow{2}{*}{$\theta_3$} & DNN-Q & -0.000 &  0.027 & 0.109 & 0.980 \\
   & GP-Fwd & 0.114 &0.233& 0.098 & 0.300\\
    \hline
    \end{tabular}
    \label{tab:simul2}
    \end{center}
    $\dagger$: Average Length of 95\% Prediction Interval.\\
    $\ddag$: Empirical Coverage of 95\% Prediction Interval.
\end{table}

In Table \ref{tab:simul2} we compare the performance of DNN-Q and GP-Fwd based on 50 selected cases out of the 1,500 test cases, since applying GP-Fwd to all 1,500 test cases is computationally too expensive due to the need of running a long MCMC chain for each case. The details on how these 50 cases are selected is described in Section \ref{sec:DataGenerationDetails}. 
As we have seen in Table \ref{tab:simul1} DNN-Q method provides accurate point estimates and sound uncertainty quantification for all three input parameters. On the contrary GP-Fwd results in overly dispersed prediction intervals for the first parameter (that cover almost the entire parameter range $[0,1]$) and severe biases and undercoverage of the prediction intervals for the second and third parameters.
 If we know the discrepancy parameters with a high confidence and impose strong priors accordingly GP-Fwd may  suffer less from inferential issues but assuming that the form of discrepancy is exactly known is highly unrealistic in practice. 
Another important limitation of GP-Fwd is the difficulty of building an accurate emulator. The emulation performance evaluation described in Section \ref{sec:emulation} shows that the Gaussian process emulator does not provide a satisfactory prediction accuracy in this emulation problem. One might be able to improve the emulation performance by incorporating a more complicated (and potentially non-stationary) dependence structure in the Gaussian process emulator model, but such added complexity may cause computational and inferential challenges.

\section{Application to WRF-Hydro Model} \label{sec:application}

In this section we apply our proposed DNN-Q method to the problem of calibrating WRF-Hydro \citep{gochis2018wrf}, the hydrologic extension of WRF model to demonstrate that our method can be used to calibrate a highly complicated computer model and provide useful information about the input parameter uncertainty and the data-model discrepancy. 
The WRF-Hydro model provides an innovative way to simulate the entire water cycle (surface and sub-surface runoff, and channel routing) by coupling a land surface component and high-resolution hydrologic components. It contains a large number of uncertain parameters that need to be properly tuned for realistic simulations \citep{wang2019parallel}. 

\begin{figure}[h]
\centering
\includegraphics[scale=0.6]{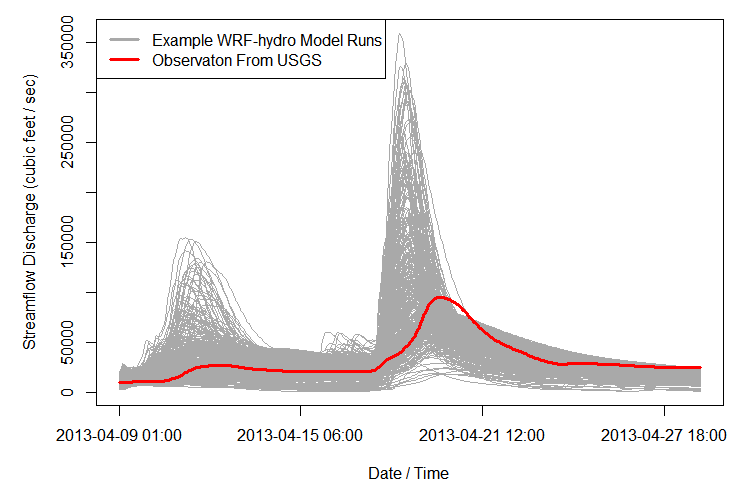}
\caption{WRF-hydro model outputs and observational data for Iowa River at Wapello, IA (USGS ID\#05465500)}
\label{fig:ModelRunsAndObs}
\end{figure}

The observational data are collected by the United States Geological Survey (USGS). The objective is to find the best input parameter setting for simulating the streamflow at Iowa River at Wapello, IA (USGS ID\#05465500), and the relevant model output and observational data are time series for the same time period. The model ensemble has 400 members with 14 varied parameters but we calibrate only five of them as the other parameters are not relevant to the terrain types of the target site.  We provide more detailed description about the input parameters in Section \ref{sec:application_details}. The simulated and observed time series are the average water volume (feet$^3$/sec) of streamflow for 15 minutes intervals recorded from April 9th to 28th in 2013, having 480 time steps in total. This period had a major precipitation event in the area and hence provides useful information on input parameters relevant to modeling streamflows. The model runs and observational data are shown in Figure \ref{fig:ModelRunsAndObs}.

As shown in Figure \ref{fig:ModelRunsAndObs} the observational data do not resemble any of the model runs, suggesting that there are some notable data-model discrepancies. This suggests that our inverse model-based approach is useful to properly estimate the parameter values while accounting for the data-model discrepancy in this problem. The range, partial sill, and nugget parameters for the discrepancy term $\bdelta$ are sampled from an improved Latin hypercube design. The sampling range for the discrepancy range parameter ($\phi$) is set to be [10,300] as in the simulation study in Section \ref{sec:simulation_study} to train our DNN model based on discrepancy patterns with various time scales. The range for the nugget  parameter ($\zeta$) set to be [1,10] to reflect the fact that both the model output and the observational data show very smooth trends. To determine the range for the partial sill ($\kappa$) we have conducted some exploratory data analysis and found that the interquartile range for the mean squared error between the model output and the observational data range from 10,292 (ft$^3$/sec) to 21,670 (ft$^3$/sec). Loosely based on this observation we set the range for the partial sill parameter to be $[5000^2,15000^2]$ so that the lower bound is well less than 10,292 and $2\times$(the upper bound)=30,000 well exceeds 21,670.

\begin{figure}[h]
\centering
\includegraphics[scale=0.6]{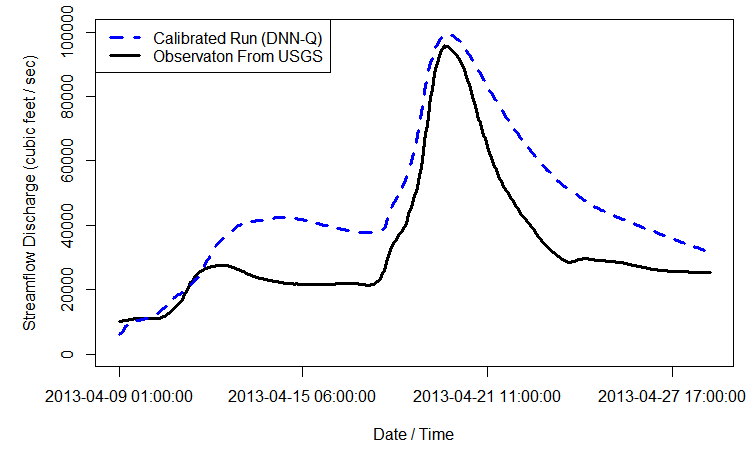}
\caption{WRF model runs at the estimated parameter settings by DDN-Q (blue) compared to the observations from USGS (black)}
\label{fig:CalibratedRun}
\end{figure}

The estimated parameter values based on the observational data are summarized in Table \ref{tab:obs}. Using the median estimates for the parameter we run WRF-Hydro model and compared the simulated streamflow with the observational data from USGS (Figure \ref{fig:CalibratedRun}). Compared to the all model runs in the ensemble shown in Figure \ref{fig:ModelRunsAndObs} the parametric uncertainty in simulation is significantly reduced. The calibrated run have accurately captured two important hydrologic quantities, the timing and the magnitude of the peak stream flow discharge (with a slight overestimation for the magnitude, though). Note however there are notable discrepancies before and after the peak surges, which may be due to WRF-Hydro model deficiencies in capturing certain hydrological processes that need to be improved or taken into account.

\section{Summary and Future Directions}

In this paper we have proposed a new computer model calibration method using deep learning. The framework focuses on the case where the model output and observational data are in the form of time series but the basic framework can be easily modified for other types of data such as spatial data or spatio-temporal data by substituting the LSTM feature extraction layers with convolutional layers \citep[e.g.][Chapter 9]{goodfellow2016deep} or convolutional LSTM layers \citep{shi2015convolutional}. Utilizing the feature extraction capacity of LSTM layers and the flexibility of fully-connected layers our DNN-based method provides an accurate way to capture the inverse relationship between the model output and the input parameters. Using `learning with noise' idea we train a DNN emulator for inverse relationship that can efficiently filter out the effects from data-model discrepancy on input parameter estimation. This provides a viable solution to one of the long-standing issues in computer model calibration literature, non-identifiablity between the effects of input parameters and data-model discrepancy. Our framework also provides a way to quantify the uncertainty in parameter estimation in the form of interval estimates using quantile regression. This approach can be used to quantify the uncertainty in any DNN-based modeling problems and hence has an implication beyond the problem of computer model calibration.

As per possible future extensions one possible direction is to modify our framework so that it can handle non-continuous data such as binary or count data. This requires generating non-continuous contaminated model outputs and hence some generalized linear model-type approach is needed. Another possible extension is to formulate a DNN-based calibration method for temporally or spatially varying input parameters, which will require handling of high-dimensional response variables with complicated dependence structures in DNN modeling. 
All of these possible future developments have to be accompanied with  development of a proper uncertainty quantification method through a statistical inference procedure that is specifically designed for particular distributional assumptions and variable types at hand.

\if0\blind
{
\section*{Acknowledgment}

This research is partially supported by the University of Cincinnati Charles Phelps Taft Research Center and the Ohio Super Computing Center (OSC).
} \fi

\bibliographystyle{Chicago.bst}

\bibliography{long,references}
\end{document}